\documentclass[letterpaper, 10 pt, conference]{ieeeconf}

\IEEEoverridecommandlockouts
\usepackage{cite}
\usepackage{amsmath,amssymb,amsfonts}
\usepackage{algorithmic}
\usepackage[short]{optidef}
\usepackage{physics}
\usepackage{graphicx}
\usepackage{textcomp}
\usepackage{xcolor}
\usepackage{hyperref}

\usepackage{bm}

\usepackage{multirow}
\usepackage{booktabs}
\usepackage{threeparttable}
\usepackage{makecell}

\usepackage{tikz}
\usepackage[utf8]{inputenc}
\usepackage{pgfplots} 
\usepackage{pgfgantt}
\usepackage{pdflscape}
\pgfplotsset{compat=newest} 
\pgfplotsset{plot coordinates/math parser=false}

\def\*#1{\bm{#1}}

\begin{document}

\title{\LARGE \bf MCOO-SLAM: A Multi-Camera Omnidirectional Object SLAM System}
\author{Miaoxin Pan, Jinnan Li, Yaowen Zhang, Yi Yang$^{*}$, Yufeng Yue
\thanks{This work was partly supported by National Natural Science Foundation of China (Grant No. NSFC 62233002) and National Key R\&D Program of China (2022YFC2603600).}
\thanks{All authors are with the School of Automation, Beijing Institute of Technology, Beijing 100081, China.}
}

\maketitle
\thispagestyle{empty}
\pagestyle{empty}

\begin{abstract}
Object-level SLAM offers structured and semantically meaningful environment representations, making it more interpretable and suitable for high-level robotic tasks. However, most existing approaches rely on RGB-D sensors or monocular views, which suffer from narrow fields of view, occlusion sensitivity, and limited depth perception—especially in large-scale or outdoor environments. These limitations often restrict the system to observing only partial views of objects from limited perspectives, leading to inaccurate object modeling and unreliable data association. In this work, we propose MCOO-SLAM, a novel Multi-Camera Omnidirectional Object SLAM system that fully leverages surround-view camera configurations to achieve robust, consistent, and semantically enriched mapping in complex outdoor scenarios. Our approach integrates point features and object-level landmarks enhanced with open-vocabulary semantics. A semantic-geometric-temporal fusion strategy is introduced for robust object association across multiple views, leading to improved consistency and accurate object modeling, and an omnidirectional loop closure module is designed to enable viewpoint-invariant place recognition using scene-level descriptors. Furthermore, the constructed map is abstracted into a hierarchical 3D scene graph to support downstream reasoning tasks. Extensive experiments in real-world demonstrate that MCOO-SLAM achieves accurate localization and scalable object-level mapping with improved robustness to occlusion, pose variation, and environmental complexity.
\end{abstract}

\section{Introduction}
Simultaneous Localization and Mapping (SLAM) is a foundational capability in robotics. Cameras, being lightweight and rich in geometric and semantic information, have become widely adopted sensors for SLAM across applications\cite{yang2021towards,cheng2022review,pan2024robust}. While traditional visual SLAM systems primarily rely on low-level point features, they often struggle to provide structured and interpretable representations of complex environments. This limitation has motivated growing interest in object-level SLAM, which models the world using semantically meaningful and geometrically coherent entities—such as cuboids or dual quadrics—to support structured mapping and downstream tasks like object retrieval and semantic navigation\cite{yang2019cubeslam, rubino20173d, zins2022oa}.

Despite promising advances, existing object-level SLAM systems still face critical limitations. Many of them rely on monocular RGB-D sensors, which suffer from sensitivity to lighting conditions and limited sensing range, making them unsuitable for outdoor or large-scale scenarios. Narrow fields of view and frequent occlusions in cluttered scenes also hinder consistent object association across time and viewpoints. The resulting partial and fragmented object observations often lead to incomplete or inaccurate models, limiting the robustness and scalability of these systems in real-world environments that demand long-term autonomy and environmental diversity.

\begin{figure}
	\vspace{2mm}
	\centering
	\includegraphics[width=\linewidth]{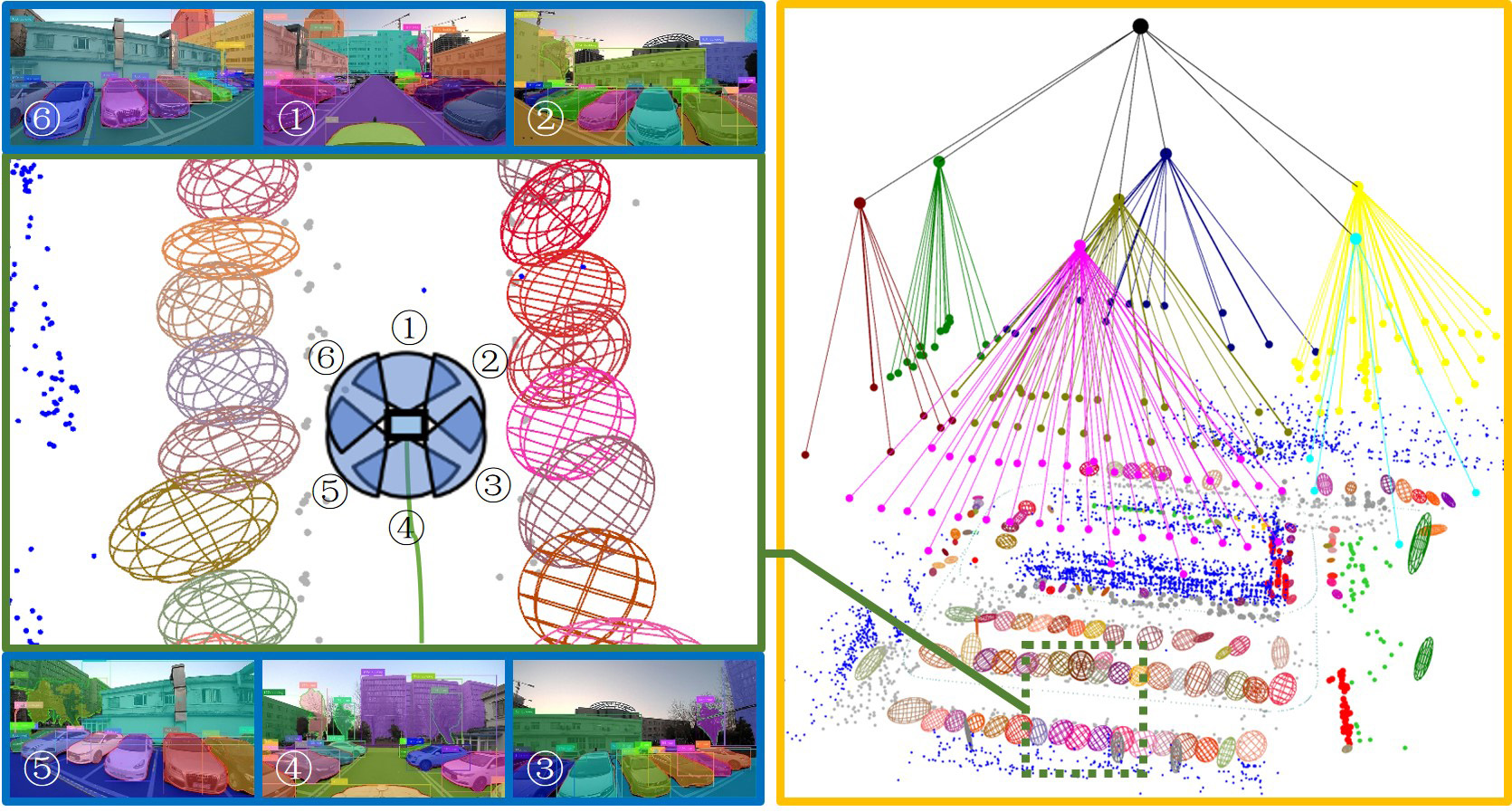}
	\vspace{-7mm}
	\caption{MCOO-SLAM takes omnidirectional multi-camera images as input and fuses low-level geometric features and high-level open-vocabulary semantics, enabling robust multi-view object association, omnidirectional loop closure, and hierarchical scene understanding for complex outdoor environments.}
	\label{fig:summary}
	\vspace{-4.5mm}
\end{figure}

To overcome these limitations, multi-camera SLAM has emerged as a promising direction. By equipping robots with surround-view cameras, systems benefit from omnidirectional perception, stronger spatial constraints, and improved robustness to occlusions\cite{yang2024mcov, yang2020multi}. These advantages naturally suggest that integrating multi-camera configurations into object-level SLAM could lead to more comprehensive and reliable scene understanding. However, this integration remains non-trivial and underexplored challenges. First, although surround-view setups provide broader coverage, object observations are often fragmented across different cameras with varying viewpoints, making it difficult to reconstruct coherent and complete object models. This requires a unified observation and optimization strategy that can jointly reason over multi-camera inputs. Second, the appearance and geometry of objects may vary significantly across views and time, and occlusions or semantic ambiguity further complicate the task of accurately associating object observations. A robust multi-camera object association method is needed to maintain consistent tracking for the surround-view system. Third, while omnidirectional perception enhances spatial awareness, current systems lack global, semantically meaningful representations for robust loop closure and high-level reasoning. There is a need for descriptors and map structures that support viewpoint-invariant recognition and structured semantic understanding.

In this article, we propose \textbf{MCOO-SLAM}, a novel \textbf{Multi-Camera Omnidirectional Object SLAM} framework designed specifically for multi-camera systems in outdoor environments. Our system integrates low-level feature points and high-level object landmarks, enhanced with semantic attributes, and constructs a hierarchical 3D scene graph to support high-level reasoning. To the best of our knowledge, this is the first object-centric SLAM system that leverages the full potential of surround-view perception. Specifically, we design dedicated object observation models and optimization strategies for surround-view systems. By leveraging the omnidirectional sensing capabilities, we propose a multi-camera object-level data association method based on semantic, geometric, and temporal fusion, thereby improving the accuracy of object modeling. During the mapping process, open-vocabulary semantic attributes are embedded into both point-level and object-level landmarks, enabling flexible object retrieval. Furthermore, we design an omnidirectional scene descriptor to better capture the surrounding environment and improve the loop closure performance. Finally, we abstract the constructed map into a hierarchical 3D scene graph, facilitating downstream high-level tasks such as querying or reasoning.

In summary, this article makes the following contributions:
\begin{itemize}
	\item A multi-camera object-level data association method that fuses semantic, geometric, and temporal cues, tailored for wide field-of-view and surround-camera systems, to achieve consistent and accurate tracking across different views and time.
	\item An omnidirectional loop closure module that constructs global scene descriptors enriched with open-vocabulary semantic information, achieving robust place recognition under large pose variations.
	\item An integrated SLAM framework that unifies feature- and object-level components, integrates the proposed modules, and supports hierarchical 3D scene graph construction. Experimental results verify the overall system’s performance and its applicability to real-world scenarios.
\end{itemize}

The remainder of this paper is organized as follows: Section \ref{sec:relatedwork} reviews related work; Section \ref{sec:MCOO-SLAM} presents the overall system architecture and detailed descriptions of the proposed methodology.
\begin{figure*}[ht]
	\centering
	\includegraphics[width=\textwidth]{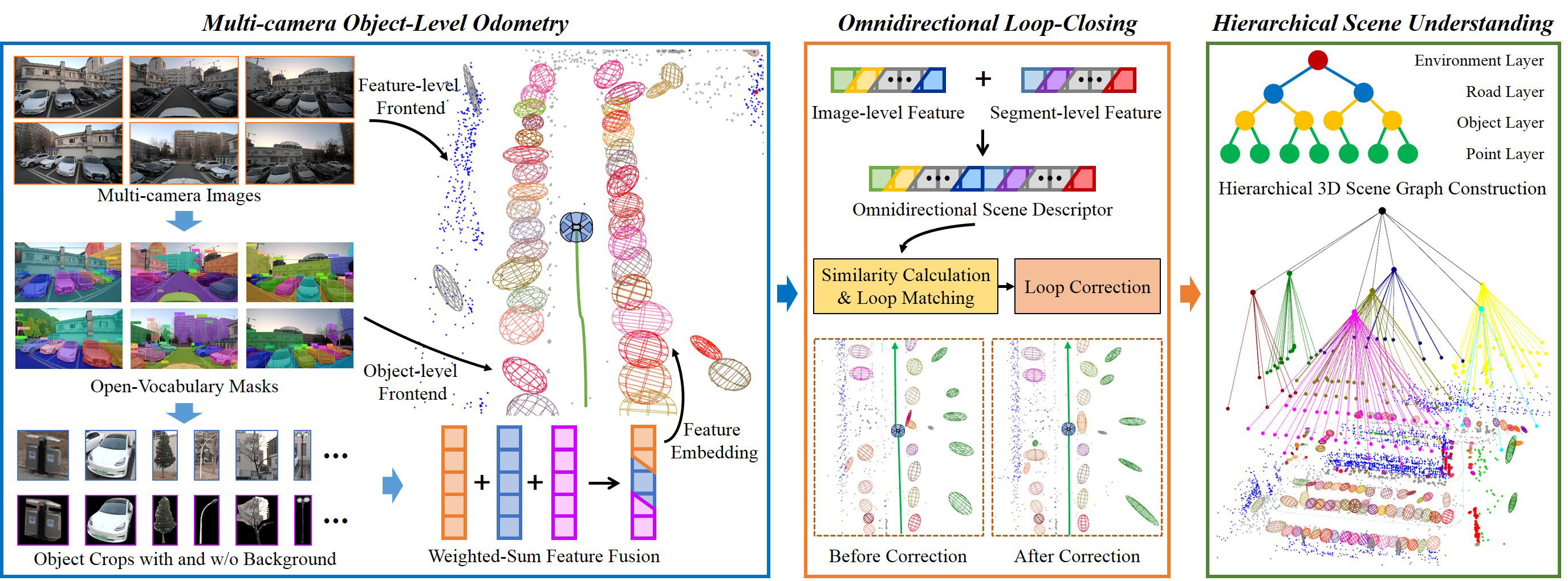}
	\vspace{-6mm}
	\caption{Framework of the proposed MCOO-SLAM system, which mainly consists of three parts: multi-camera object-level odometry, omnidirectional loop-closing, and hierarchical scene understanding.}
	\vspace{-6mm}
	\label{fig:systemoverview}
\end{figure*}
\section{Related Work}\label{sec:relatedwork}
\subsection{Multi-camera Visual SLAM}\label{subsec:mcvslam}
To expand the perceptual field and improve robustness, researchers have extended traditional visual SLAM to multi-camera configurations. Early systems like MCPTAM\cite{harmat2012parallel} improved tracking resilience through parallel multi-view processing, and later works further enhanced it by incorporating multi-camera extrinsic calibration\cite{harmat2015multi} and improving scale recovery capabilities\cite{tribou2015multi}. To address large-scale and unstructured outdoor scenes, Yang et al.\cite{yang2020multi} proposed a collaborative multi-camera SLAM framework suited for off-road scenarios. Seok et al.\cite{seok2019rovo} introduced a hybrid projection model for visual odometry, achieving robust performance in open environments using optical flow. Similarly, 360VO~\cite{Huang2022360vo} leveraged 360° surround-view camera and minimized photometric error to reconstruct denser maps than conventional indirect methods. Multicol-SLAM\cite{urban2016multicol} initialized maps using overlapping views but did not exploit cross-camera constraints during runtime, leading to potential scale drift. PAN-SLAM~\cite{Ji2020panslam} unified multi-fisheye views into a spherical domain for consistent mapping, but lacked baseline constraints and efficient loop detection. MCOV-SLAM\cite{yang2024mcov} further improved omnidirectional robustness by fusing geometric cues from all cameras to enable place recognition under arbitrary orientations. While these systems improve geometric robustness and perceptual coverage, they mainly remain feature-based and do not utilize semantic understanding. This limits their applicability to high-level tasks such as semantic navigation or object-based retrieval, leaving room for improvement through the integration of semantic, especially open-vocabulary, understanding into multi-camera SLAM.

\subsection{Object-oriented SLAM}\label{subsec:ooslam}
To address the limitations of low-level feature maps, object-oriented SLAM introduces high-level semantic entities into the SLAM pipeline, offering more structured, compact, and interpretable representations of the environment. Early works like SLAM++\cite{salas2013slam} achieved efficient object mapping using known 3D models, but lacked flexibility for open-world deployments. Later systems such as Fusion++\cite{mccormac2018fusion} and the approach in\cite{sunderhauf2017meaningful} relaxed model assumptions using instance segmentation and volumetric fusion.To reduce model complexity, several methods adopted geometric primitives to represent objects. CubeSLAM~\cite{yang2019cubeslam} and QuadricSLAM~\cite{nicholson2018quadricslam} used cuboids and dual quadrics, respectively, trading fine-grained surfaces for efficient and compact representations. Dual quadrics, in particular, benefit from a closed-form formulation in projective geometry~\cite{rubino20173d}, enabling efficient optimization. Extensions such as~\cite{lin2021robust,liao2022so,wu2023object,wang2023qiso} have enhanced these representations by designing more precise observation models or incorporating them directly into bundle adjustment. However, the overall improvement compared to point-based SLAM remains marginal in many real-world settings. Several hybrid systems have demonstrated the complementary nature of feature points and object-level semantics. OA-SLAM~\cite{zins2022oa} first localizes coarsely using object detections and refines poses with traditional features. VOOM~\cite{wang2024voom} tightly integrates object landmarks to guide point-based associations, allowing more stable odometry through both temporal and spatial correspondence. These approaches have significantly advanced object-level SLAM by combining semantics with geometry. However, most object-oriented SLAM approaches rely on monocular or RGB-D sensors, which are constrained by limited fields of view, short sensing range, and sensitivity to lighting. These limitations hinder their scalability to large-scale outdoor deployments. 

Our work bridges these directions. We propose MCOO-SLAM, a novel multi-camera SLAM framework that integrates high-level object semantics with low-level geometric features. Our method enables robust multi-view object association, omnidirectional loop closure via semantic scene descriptors, and hierarchical scene graph construction for downstream tasks.

\section{MCOO-SLAM}\label{sec:MCOO-SLAM}
\subsection{System Overview}\label{subsec:system_overview}
The proposed MCOO-SLAM takes omnidirectional multi-camera images as input and produces a semantically enriched, globally consistent map comprising sparse point clouds, object ellipsoids, and a hierarchical scene graph. The system architecture is divided into three main modules: Multi-camera Object-Level Odometry, Omnidirectional Loop-Closing, and Hierarchical Scene Understanding, as illustrated in Fig.\ref{fig:systemoverview}. In the front end, feature points and pose estimation are processed following the method in \cite{yang2024mcov}, while object instances are detected using Grounding DINO\cite{liu2024grounding} and segmented with SAM2~\cite{Kirillov_2023_ICCV, ravi2024sam2}. Object associations across time and views are established through a multi-level fusion of semantic, geometric, and temporal cues, using a weighted Wasserstein distance metric to enhance robustness (Sec.\ref{subsec:oom_for_mcos} and \ref{subsec:oola}). Once a keyframe is inserted, we project its segmentation masks into 3D space, encode them with CLIP~\cite{pmlr-v139-radford21a}, and cluster the features to construct a segment-level open-vocabulary map (Sec.\ref{subsec:3dovmapping}).The back end performs loop closure detection using omnidirectional open-vocabulary scene descriptors to enable viewpoint-invariant matching, followed by global optimization to refine poses and map entities (Sec.~\ref{subsec:ovolcm}).Finally, the reconstructed point clouds and object landmarks are abstracted into a hierarchical 3D scene graph, enabling high-level reasoning and semantic queries in complex outdoor environments(Sec.~\ref{subsec:h3dsgc}).

\subsection{Object Observation Model for Multi-camera Systems}\label{subsec:oom_for_mcos}
Most object observation models in SLAM assume a pinhole camera and ignore lens distortion, enabling a simplified linear projection from 3D dual quadrics to 2D ellipses. However, this assumption breaks down for wide-angle or fisheye cameras in surround-view systems, where non-linear projection models (e.g., CMEI, unified omnidirectional models) are essential due to significant image distortion. Consequently, the conventional linear projection between 3D dual quadrics and 2D ellipses becomes invalid. To address this, we propose a non-linear object observation model that explicitly accounts for distortion. Instead of directly projecting the 3D dual quadric $\mathbf{Q}_k^* \in \mathbb{R}^{4\times4}$ onto the distorted image plane, we first back-project contour points $\mathbf{u}_i = [u_i, v_i]^\top$ from instance segmentation to normalized rays via the inverse camera model $\pi^{-1}(\cdot)$:
\vspace{-1mm}
\begin{equation}
	\mathbf{p}_i^{\text{norm}} = \frac{\pi^{-1}(\mathbf{u}_i)}{\|\pi^{-1}(\mathbf{u}_i)\|_z}
\end{equation}
\vspace{-1mm}

The resulting normalized points set $\{\mathbf{p}_i^{\text{norm}}\}$ is used to fit a dual conic $\mathbf{C}_{fk}^{*\text{norm}} \in \mathbb{R}^{3\times3}$ on the normalized plane, establishing the projection relationship:
\vspace{-1mm}
\begin{equation}
	\mathbf{C}_{fk}^{*\text{norm}} = \mathbf{P}_f^{\text{norm}} \mathbf{Q}_k^* \mathbf{P}_f^{\text{norm}^\top}
\end{equation}
\vspace{-1mm}
where $\mathbf{P}_f^{\text{norm}} = [\mathbf{R}_f | \mathbf{t}_f]$ is the camera projection matrix in normalized space.

To quantify the consistency between predicted and observed ellipses, we adopt a Gaussian representation of ellipses:
\vspace{-1mm}
\begin{equation}
	\boldsymbol{\mu} = 
	\begin{pmatrix}
		p_x^{\text{norm}} \\
		p_y^{\text{norm}}
	\end{pmatrix}, \quad
	\boldsymbol{\Sigma}^{-1} = \mathbf{R}^\top(\theta)
	\begin{pmatrix}
		\frac{1}{\alpha^2} & 0 \\
		0 & \frac{1}{\beta^2}
	\end{pmatrix}
	\mathbf{R}(\theta)
\end{equation}
\vspace{-1mm}
And the second-order Wasserstein distance on the normalized plane serves as our observation residual:
\vspace{-1mm}
\begin{equation}
	\mathcal{W}_2^2(\mathcal{N}_{fk}^{\text{obs}}, \mathcal{N}_{fk}^{\text{est}}) =
	\|\boldsymbol{\mu}_1 - \boldsymbol{\mu}_2\|_2^2 + \|\boldsymbol{\Sigma}_1^{1/2} - \boldsymbol{\Sigma}_2^{1/2}\|_F^2
\end{equation}
\vspace{-1mm}
where $\|\cdot\|_F$ denotes the Frobenius norm.

Unlike monocular systems that rely on sparse, temporally separated observations—which are prone to occlusion and partial contours—we introduce a joint multi-camera, multi-frame estimation strategy. Let $\mathcal{C}$ be the set of cameras and $\mathcal{T}$ the set of keyframes. The full set of spatiotemporal observations is:
\vspace{-1mm}
\begin{equation}
	\mathcal{F}_L = \bigcup_{t \in \mathcal{T}} \bigcup_{c \in \mathcal{C}} \{(c, t)\}
\end{equation}
\vspace{-1mm}

We jointly optimize the dual quadric $\mathbf{Q}_k^*$ over all observed ellipses across $\mathcal{F}_L$ using the total Wasserstein residual:
\vspace{-1mm}
\begin{equation}
	e_{\text{proj}} = \sum_{(c, t) \in \mathcal{F}_L} 
	\mathcal{W}_2^2(\mathcal{N}_{ctk}^{\text{obs}}, \mathcal{N}_{ctk}^{\text{est}})
\end{equation}

To improve geometric stability under partial visibility, we introduce an additional residual on the object center. Let $\mathcal{P}_k$ be the associated 3D points, and $\mathbf{C}^{\text{obs}}$ their centroid. We define:
\vspace{-1mm}
\begin{equation}
	e_{\text{center}} = \mathbf{C}^{\text{est}} - \mathbf{C}^{\text{obs}}
\end{equation}
\vspace{-1mm}
This residual is weighted by the number of observations and incorporated into the optimization graph.

The final objective jointly minimizes both residuals:
\vspace{-1mm}
\begin{equation}
	\arg\min_{\mathbf{Q}_k^*} \sum_{(c, t) \in \mathcal{F}_L} 
	\mathcal{W}_2^2\left( \mathcal{N}_{ctk}^{\text{obs}}, \mathcal{N}_{ctk}^{\text{est}} \right)
	+ w_c \cdot \left\| \mathbf{C}^{\text{est}} - \mathbf{C}^{\text{obs}} \right\|^2
\end{equation}
\vspace{-1mm}
where $w_c$ regulates the influence of the center constraint.

\subsection{Multi-level Omnidirectional Object data Association}\label{subsec:oola}
Existing object-oriented SLAM methods based on dual quadric typically adopt two object association strategies: (1) IoU matching between reprojected and detected bounding boxes, or (2) Wasserstein distance-based residuals. While effective in small-scale or indoor settings, these approaches struggle in large-scale outdoor environments, particularly for cross-frame and cross-view association in multi-camera systems. As a result, they often suffer from mismatches and fail to meet the consistency demands of omnidirectional SLAM.

To address these challenges, we propose a three-stage object association framework that integrates semantic, geometric, and temporal cues. It establishes a consistent indexing mechanism across cameras and timestamps. By leveraging the perceptual capabilities of Grounding DINO and SAM2, combined with geometric consistency validation and historical trajectory maintenance, the proposed method significantly enhances robustness and continuity in object tracking.

We first utilize Grounding DINO to extract semantic categories and instance-level object proposals. Then, SAM2 is employed for fine-grained instance segmentation and cross-frame matching. For each single camera sequence, SAM2 provides semantic-consistent tracking across timestamps, forming reliable cross-frame object trajectories. Simultaneously, we treat observations from different cameras at the same timestamp as a pseudo-temporal sequence, enabling SAM2’s matching capability to operate across views. This cross-view association unifies object instances seen from different viewpoints. Finally, by synchronizing timestamps and matching semantic identities, we fuse cross-frame and cross-view associations into a unified object index:
\vspace{-1mm}
\begin{equation}
	o_{c,t,k} \longleftrightarrow o_{c',t',k}, \quad \forall~c, c' \in \mathcal{C},~t, t' \in \mathcal{T}
\end{equation}
\vspace{-1mm}
where $o_{c,t,k}$ denotes the $k$-th object observed by the $c$-th camera at time $t$.

Although SAM2 exhibits strong matching capability, errors such as drift or misassociation can occur in crowded or ambiguous scenes. To address this, we introduce a geometric consistency check based on a novel weighted Wasserstein distance formulation. In practice, we observe that the traditional Wasserstein distance may yield deceptively small values when two ellipses have similar axis lengths and orientations but are spatially separated with minimal or no overlap. This often results in incorrect associations in large-scale scenes. To mitigate this, we propose a spatially-weighted matching score that combines both shape similarity and overlap extent:
\vspace{-1mm}
\begin{equation}
	W_{\text{score}}(e_{fk}^{\text{est}}, e_{fk}^{\text{obs}}) = 
	\text{IoU}(e_{fk}^{\text{est}}, e_{fk}^{\text{obs}}) \cdot \exp\left( -\frac{\sqrt{\mathcal{W}_2^2(\mathcal{N}_1, \mathcal{N}_2)}}{C} \right)
\end{equation}
\vspace{-1mm}
where $\text{IoU}(\cdot, \cdot)$ denotes the Intersection over Union between the estimated ellipse $e_{fk}^{\text{est}}$ and the detected ellipse $e_{fk}^{\text{obs}}$, and $\mathcal{W}_2^2$ denotes the second-order Wasserstein distance between their Gaussian representations. This formulation penalizes geometric dissimilarity and poor spatial overlap simultaneously, effectively filtering out false positives caused by spatial displacement.

To maintain temporal coherence, we introduce a dynamic object-level memory mechanism. For each object successfully associated across views or frames, we record its spatiotemporal observations in a trajectory index table. During future association steps, the system first queries this memory for potential candidates, enabling rapid re-matching and bypassing redundant semantic/geometric checks. This memory supports the re-identification of temporarily occluded or re-entering objects and enhances continuity in long-term tracking.

By sequentially integrating semantic detection, geometric validation, and temporal memory, our association method significantly improves accuracy and robustness in surround-view SLAM, particularly under clutter and large-scale multi-view conditions. The resulting object associations are subsequently utilized to enhance the matching and optimization of feature points, similar to the strategy employed in \cite{wang2024voom}, thereby contributing to enhanced pose estimation accuracy.

\subsection{3D Segment-Level Open-Vocabulary Mapping}\label{subsec:3dovmapping}
Building upon the previously established map points and object-level instances from our multi-camera SLAM system, we construct an open-vocabulary semantic map at the segment level by associating vision-language features with 3D data. This mapping enables more expressive and flexible semantic understanding of the environment, moving beyond traditional closed-set categorizations.

During the earlier image processing stage, class-agnostic 2D segmentation results were generated. In this step, we associate those segmentation masks with map points and objects. Inspired by \cite{werby23hovsg}, each 2D mask is encoded using CLIP from three perspectives: full image, masked region with background, and masked region without background. The resulting features are fused via a weighted sum to balance global context and object-level detail. For each 3D segment, we collect the features of all associated 2D masks across views. DBSCAN is applied to cluster these features and suppress outliers. The segment is then assigned the feature vector nearest to the dominant cluster centroid, providing a robust and meaningful open-vocabulary label.

\subsection{Omnidirectional Loop-closing Method}\label{subsec:ovolcm}
Loop-closing plays a vital role in SLAM systems by correcting accumulated drift through recognizing previously visited places and introducing additional pose constraints. However, traditional loop-closing approaches based on local visual features (e.g., keypoints and descriptors) often struggle under varying viewpoints or illumination conditions, limiting their robustness in complex real-world environments.

To address these issues, we propose an omnidirectional scene-level loop-closing framework that combines high-level semantic understanding with the wide field-of-view provided by surround-view cameras. By leveraging open-vocabulary semantic cues and omnidirectional perception, our method improves both robustness and accuracy in loop detection and correction.

\begin{figure}[!h]
	\centering
	\includegraphics[width=\columnwidth]{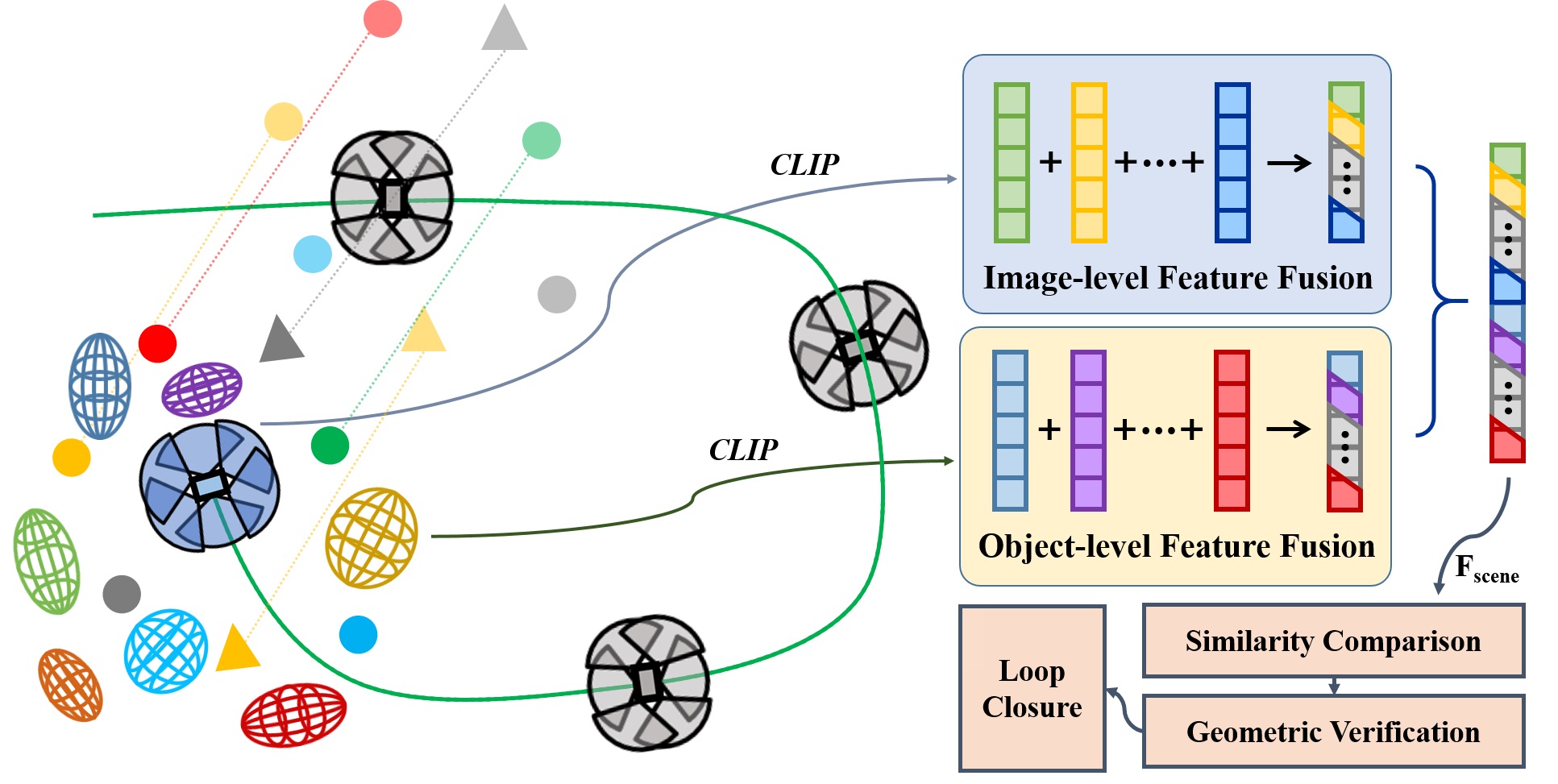}
	\caption{Overview of the Omnidirectional Open-vocabulary Loop Closure Detection Framework. Each keyframe is encoded into a hierarchical scene descriptor, combining global image-level semantics and segment-level semantics. This approach enables viewpoint-invariant and semantically robust loop detection across omnidirectional multi-camera inputs.}
	\label{fig:lcd}
\end{figure}
\vspace{-3mm}
\subsubsection{Omnidirectional loop closure detection}\label{subsubsec:ovolcd}
For each keyframe, we construct a hierarchical descriptor $\textbf{F}_\text{scene} = [\textbf{F}_\text{view}; \textbf{F}_\text{seg}]$, referred to as the omnidirectional open-vocabulary scene descriptor, to summarize the surrounding scene semantically at both the image and segment levels. At the image level, CLIP features from all camera views are averaged and normalized to yield $\textbf{F}_\text{view}$. At the segment level, features $\textbf{f}_{\text{seg}_i}$ from objects or clusters are aggregated using weights $w_i$ inversely proportional to their distances from the ego system, resulting in $\textbf{F}_\text{seg}$. The formulation is:
\begin{equation}\label{eq:global_feature}
	\textbf{F}_\text{view} = \frac{1}{N_{\text{cams}}} \sum_{i=1}^{N_{\text{cams}}} \textbf{f}_{\text{cam}_i}, \quad
	\textbf{F}_\text{seg} = \sum_{i=1}^{N_{\text{segments}}} w_i \textbf{f}_{\text{seg}_i}
\end{equation}

To detect loops, we compute a similarity score between the current keyframe and prior candidates, as Equations (\ref{eq:similarity_calc}), where the superscripts $c$ and $h$ represent the current keyframe and the historical keyframe respectively. Specifically, we evaluate the similarity between current keyframe and other keyframes with which it co-views no less than a certain number of map points, and use the lowest similarity as the threshold to screen all keyframes to find putative matches. After we have a putative loop closure between our query and match keyframes, we attempt to compute a relative pose between the two by performing a RANSAC-based geometric verification as in \cite{yang2024mcov}. Thanks to the integration of omnidirectional imagery and semantic descriptors, our method achieves viewpoint-invariant loop detection, robust even to large orientation changes.
See Fig. \ref{fig:lcd} for a visual summary.
\vspace{-1mm}
\begin{equation}\label{eq:similarity_calc}
	S_{\text{total}} = \alpha \cdot \cos(\textbf{F}_\text{view}^{(\text{c})},\textbf{F}_\text{view}^{(\text{h})}) + \beta \cdot \cos(\textbf{F}_\text{seg}^{(\text{c})},\textbf{F}_\text{seg}^{(\text{h})})
\end{equation}

\subsubsection{Omnidirectional loop closure correction}\label{subsubsec:ovolcc}
Once the loop closures are detected, we integrate these constraints into the global SLAM back-end, jointly optimizing both pose and map consistency. In addition to the correction of map points, we introduce an object-level loop closure optimization mechanism. When a keyframe’s pose $\mathbf{T}_k$ is updated to $\mathbf{T}_k^{\text{opt}}$, we utilize the elliptical object observations detected in this keyframe, together with the optimized pose, to construct a residual term for the optimization of the ellipsoid representation $\mathbf{Q}_{obj}^*$. After optimization, we further merge redundant object instances based on spatial proximity and semantic similarity, thereby enhancing the geometric representation and spatial consistency of object entities within the map.

\subsection{Hierarchical 3D Scene Graph Construction}\label{subsec:h3dsgc}
To support efficient semantic organization and high-level reasoning in complex outdoor environments, we construct a hierarchical 3D scene graph grounded in object-level information and point clouds derived from our multi-camera SLAM system. Inspired by recent advances such as OpenGraph~\cite{deng2024opengraph}, our scene graph is structured across multiple layers of abstraction to capture both geometric and semantic relationships:(1) the root layer represents the entire environment; (2) the road layer includes structural elements such as roads and intersections; (3) the object layer comprises fused object observations aggregated across views and time; (4) the final layer encodes the optimized 3D map, including the pose graph and semantically annotated point clouds. This hierarchical design mirrors the compositional structure of real-world environments and enables interaction, navigation, and semantic reasoning at multiple levels of detail.

By bridging low-level perception with high-level semantics, the constructed scene graph provides an interpretable, multi-resolution representation that facilitates downstream tasks such as object retrieval, open-vocabulary querying, and context-aware planning in outdoor settings.

\vspace{-1mm}

\bibliographystyle{IEEEtran}
\bibliography{ref}

\begin{thebibliography}{10}
\providecommand{\url}[1]{#1}
\csname url@samestyle\endcsname
\providecommand{\newblock}{\relax}
\providecommand{\bibinfo}[2]{#2}
\providecommand{\BIBentrySTDinterwordspacing}{\spaceskip=0pt\relax}
\providecommand{\BIBentryALTinterwordstretchfactor}{4}
\providecommand{\BIBentryALTinterwordspacing}{\spaceskip=\fontdimen2\font plus
\BIBentryALTinterwordstretchfactor\fontdimen3\font minus
  \fontdimen4\font\relax}
\providecommand{\BIBforeignlanguage}[2]{{%
\expandafter\ifx\csname l@#1\endcsname\relax
\typeout{** WARNING: IEEEtran.bst: No hyphenation pattern has been}%
\typeout{** loaded for the language `#1'. Using the pattern for}%
\typeout{** the default language instead.}%
\else
\language=\csname l@#1\endcsname
\fi
#2}}
\providecommand{\BIBdecl}{\relax}
\BIBdecl

\bibitem{yang2021towards}
Y.~Yang, M.~Pan, S.~Jiang, J.~Wang, W.~Wang, J.~Wang, and M.~Wang, ``{Towards
  autonomous parking using vision-only sensors},'' in \emph{2021 IEEE/RSJ
  International Conference on Intelligent Robots and Systems (IROS)}.\hskip 1em
  plus 0.5em minus 0.4em\relax IEEE, 2021, pp. 2038--2044.

\bibitem{cheng2022review}
J.~Cheng, L.~Zhang, Q.~Chen, X.~Hu, and J.~Cai, ``{A review of visual SLAM
  methods for autonomous driving vehicles},'' \emph{Engineering Applications of
  Artificial Intelligence}, vol. 114, p. 104992, 2022.

\bibitem{pan2024robust}
X.~Pan, G.~Huang, Z.~Zhang, J.~Li, H.~Bao, and G.~Zhang, ``{Robust
  Collaborative Visual-Inertial SLAM for Mobile Augmented Reality},''
  \emph{IEEE Transactions on Visualization and Computer Graphics}, 2024.

\bibitem{yang2019cubeslam}
S.~Yang and S.~Scherer, ``{CubeSLAM: Monocular 3-D Object SLAM},'' \emph{IEEE
  Transactions on Robotics}, vol.~35, no.~4, pp. 925--938, 2019.

\bibitem{rubino20173d}
C.~Rubino, M.~Crocco, and A.~Del~Bue, ``{3D Object Localisation from Multi-View
  Image Detections},'' \emph{IEEE transactions on pattern analysis and machine
  intelligence}, vol.~40, no.~6, pp. 1281--1294, 2017.

\bibitem{zins2022oa}
M.~Zins, G.~Simon, and M.-O. Berger, ``{OA-SLAM: Leveraging Objects for Camera
  Relocalization in Visual SLAM},'' in \emph{2022 IEEE international symposium
  on mixed and augmented reality (ISMAR)}.\hskip 1em plus 0.5em minus
  0.4em\relax IEEE, 2022, pp. 720--728.

\bibitem{yang2024mcov}
Y.~Yang, M.~Pan, D.~Tang, T.~Wang, Y.~Yue, T.~Liu, and M.~Fu, ``{MCOV-SLAM: A
  Multicamera Omnidirectional Visual SLAM System},'' \emph{IEEE/ASME
  Transactions on Mechatronics}, vol.~29, no.~5, pp. 3556--3567, 2024.

\bibitem{yang2020multi}
Y.~Yang, D.~Tang, D.~Wang, W.~Song, J.~Wang, and M.~Fu, ``{Multi-camera visual
  SLAM for off-road navigation},'' \emph{Robotics and Autonomous Systems}, vol.
  128, p. 103505, 2020.

\bibitem{harmat2012parallel}
A.~Harmat, I.~Sharf, and M.~Trentini, ``{Parallel tracking and mapping with
  multiple cameras on an unmanned aerial vehicle},'' in \emph{International
  Conference on Intelligent Robotics and Applications}.\hskip 1em plus 0.5em
  minus 0.4em\relax Montreal, QC, Canada: Springer, 2012, pp. 421--432.

\bibitem{harmat2015multi}
A.~Harmat, M.~Trentini, and I.~Sharf, ``{Multi-Camera Tracking and Mapping for
  Unmanned Aerial Vehicles in Unstructured Environments},'' \emph{Journal of
  Intelligent and Robotic Systems: Theory and Applications}, vol.~78, no.~2, p.
  291 – 317, 2015.

\bibitem{tribou2015multi}
M.~J. Tribou, A.~Harmat, D.~W. Wang, I.~Sharf, and S.~L. Waslander,
  ``{Multi-camera parallel tracking and mapping with non-overlapping fields of
  view},'' \emph{International Journal of Robotics Research}, vol.~34, no.~12,
  pp. 1480--1500, 2015.

\bibitem{seok2019rovo}
H.~Seok and J.~Lim, ``{ROVO: Robust omnidirectional visual odometry for
  wide-baseline wide-FOV camera systems},'' in \emph{2019 International
  Conference on Robotics and Automation (ICRA)}.\hskip 1em plus 0.5em minus
  0.4em\relax Montreal, QC, Canada: IEEE, 2019, pp. 6344--6350.

\bibitem{Huang2022360vo}
H.~Huang and S.-K. Yeung, ``{360VO: Visual Odometry Using A Single 360
  Camera},'' in \emph{2022 International Conference on Robotics and Automation
  (ICRA)}, 2022, pp. 5594--5600.

\bibitem{urban2016multicol}
S.~Urban and S.~Hinz, ``{MultiCol-SLAM - A Modular Real-Time Multi-Camera SLAM
  System},'' \emph{arXiv preprint arXiv:1610.07336}, 2016.

\bibitem{Ji2020panslam}
S.~Ji, Z.~Qin, J.~Shan, and M.~Lu, ``{Panoramic SLAM from a multiple fisheye
  camera rig},'' \emph{ISPRS Journal of Photogrammetry and Remote Sensing},
  vol. 159, p. 169 – 183, 2020.

\bibitem{salas2013slam}
R.~F. Salas-Moreno, R.~A. Newcombe, H.~Strasdat, P.~H. Kelly, and A.~J.
  Davison, ``{SLAM++: Simultaneous Localisation and Mapping at the Level of
  Objects},'' in \emph{2013 IEEE Conference on Computer Vision and Pattern
  Recognition}, 2013, pp. 1352--1359.

\bibitem{mccormac2018fusion}
J.~McCormac, R.~Clark, M.~Bloesch, A.~Davison, and S.~Leutenegger, ``{Fusion++:
  Volumetric Object-Level SLAM},'' in \emph{2018 international conference on 3D
  vision (3DV)}.\hskip 1em plus 0.5em minus 0.4em\relax IEEE, 2018, pp. 32--41.

\bibitem{sunderhauf2017meaningful}
N.~S{\"u}nderhauf, T.~T. Pham, Y.~Latif, M.~Milford, and I.~Reid, ``{Meaningful
  maps with object-oriented semantic mapping},'' in \emph{2017 IEEE/RSJ
  International Conference on Intelligent Robots and Systems (IROS)}.\hskip 1em
  plus 0.5em minus 0.4em\relax IEEE, 2017, pp. 5079--5085.

\bibitem{nicholson2018quadricslam}
L.~Nicholson, M.~Milford, and N.~S{\"u}nderhauf, ``{QuadricSLAM: Dual Quadrics
  From Object Detections as Landmarks in Object-Oriented SLAM},'' \emph{IEEE
  Robotics and Automation Letters}, vol.~4, no.~1, pp. 1--8, 2018.

\bibitem{lin2021robust}
X.~Lin, Y.~Yang, L.~He, W.~Chen, Y.~Guan, and H.~Zhang, ``{Robust Improvement
  in 3D Object Landmark Inference for Semantic Mapping},'' in \emph{2021 IEEE
  International Conference on Robotics and Automation (ICRA)}.\hskip 1em plus
  0.5em minus 0.4em\relax IEEE, 2021, pp. 13\,011--13\,017.

\bibitem{liao2022so}
Z.~Liao, Y.~Hu, J.~Zhang, X.~Qi, X.~Zhang, and W.~Wang, ``{SO-SLAM: Semantic
  Object SLAM With Scale Proportional and Symmetrical Texture Constraints},''
  \emph{IEEE Robotics and Automation Letters}, vol.~7, no.~2, pp. 4008--4015,
  2022.

\bibitem{wu2023object}
Y.~Wu, Y.~Zhang, D.~Zhu, Z.~Deng, W.~Sun, X.~Chen, and J.~Zhang, ``{An Object
  SLAM Framework for Association, Mapping, and High-Level Tasks},'' \emph{IEEE
  Transactions on Robotics}, vol.~39, no.~4, pp. 2912--2932, 2023.

\bibitem{wang2023qiso}
Y.~Wang, B.~Xu, W.~Fan, and C.~Xiang, ``{QISO-SLAM: Object-Oriented SLAM Using
  Dual Quadrics as Landmarks Based on Instance Segmentation},'' \emph{IEEE
  Robotics and Automation Letters}, vol.~8, no.~4, pp. 2253--2260, 2023.

\bibitem{wang2024voom}
Y.~Wang, C.~Jiang, and X.~Chen, ``{VOOM: Robust Visual Object Odometry and
  Mapping using Hierarchical Landmarks},'' in \emph{2024 IEEE International
  Conference on Robotics and Automation (ICRA)}.\hskip 1em plus 0.5em minus
  0.4em\relax IEEE, 2024, pp. 10\,298--10\,304.

\bibitem{liu2024grounding}
S.~Liu, Z.~Zeng, T.~Ren, F.~Li, H.~Zhang, J.~Yang, Q.~Jiang, C.~Li, J.~Yang,
  H.~Su \emph{et~al.}, ``{Grounding DINO: Marrying DINO with Grounded
  Pre-training for Open-Set Object Detection},'' in \emph{European Conference
  on Computer Vision}.\hskip 1em plus 0.5em minus 0.4em\relax Springer, 2024,
  pp. 38--55.

\bibitem{Kirillov_2023_ICCV}
A.~Kirillov, E.~Mintun, N.~Ravi, H.~Mao, C.~Rolland, L.~Gustafson, T.~Xiao,
  S.~Whitehead, A.~C. Berg, W.-Y. Lo, P.~Dollar, and R.~Girshick, ``{Segment
  Anything},'' in \emph{Proceedings of the IEEE/CVF International Conference on
  Computer Vision (ICCV)}, October 2023, pp. 4015--4026.

\bibitem{ravi2024sam2}
\BIBentryALTinterwordspacing
N.~Ravi, V.~Gabeur, Y.-T. Hu, R.~Hu, C.~Ryali, T.~Ma, H.~Khedr, R.~R{\"a}dle,
  C.~Rolland, L.~Gustafson, E.~Mintun, J.~Pan, K.~V. Alwala, N.~Carion, C.-Y.
  Wu, R.~Girshick, P.~Doll{\'a}r, and C.~Feichtenhofer, ``{SAM 2: Segment
  Anything in Images and Videos},'' \emph{arXiv preprint arXiv:2408.00714},
  2024. [Online]. Available: \url{https://arxiv.org/abs/2408.00714}
\BIBentrySTDinterwordspacing

\bibitem{pmlr-v139-radford21a}
A.~Radford, J.~W. Kim, C.~Hallacy, A.~Ramesh, G.~Goh, S.~Agarwal, G.~Sastry,
  A.~Askell, P.~Mishkin, J.~Clark, G.~Krueger, and I.~Sutskever, ``{Learning
  Transferable Visual Models From Natural Language Supervision},'' in
  \emph{Proceedings of the 38th International Conference on Machine Learning},
  ser. Proceedings of Machine Learning Research, M.~Meila and T.~Zhang, Eds.,
  vol. 139.\hskip 1em plus 0.5em minus 0.4em\relax PMLR, 18--24 Jul 2021, pp.
  8748--8763.

\bibitem{werby23hovsg}
A.~Werby, C.~Huang, M.~Büchner, A.~Valada, and W.~Burgard, ``Hierarchical
  open-vocabulary 3d scene graphs for language-grounded robot navigation,''
  \emph{Robotics: Science and Systems}, 2024.

\bibitem{deng2024opengraph}
Y.~Deng, J.~Wang, J.~Zhao, X.~Tian, G.~Chen, Y.~Yang, and Y.~Yue, ``{OpenGraph:
  Open-Vocabulary Hierarchical 3D Graph Representation in Large-Scale Outdoor
  Environments},'' \emph{IEEE Robotics and Automation Letters}, vol.~9, no.~10,
  pp. 8402--8409, 2024.

\end{thebibliography}

\end{document}